# LSMVOS: Long-Short-Term similarity matching for video object segmentation

Zhang Xuerui, Yuan Xia

*Nanjing University of Science and Technology, School of Computer Science and Engineering, Jiangsu Nanjing 210094, China*

**Abstract: Objective** Semi-supervised video object segmentation refers to segmenting the object in subsequent frames given the object label in the first frame. Existing algorithms are mostly based on the objectives of matching and propagation strategies, which often make use of the previous frame with masking or optical flow. This paper explores a new propagation method, uses short-term matching modules to extract the information of the previous frame and apply it in propagation, and proposes the network of Long-Short-Term similarity matching for video object segmentation (LSMOVS) **Method:** By conducting pixel-level matching and correlation between long-term matching module and short-term matching module with the first frame and previous frame, global similarity map and local similarity map are obtained, as well as feature pattern of current frame and masking of previous frame. After two refine networks, final results are obtained through segmentation network. **Results**: According to the experiments on the two data sets DAVIS 2016 and 2017, the method of this paper achieves favorable average of region similarity and contour accuracy without online fine tuning, which achieves 86.5% and 77.4% in terms of single target and multiple targets. Besides, the count of segmented frames per second reached 21. **Conclusion:** The short-term matching module proposed in this paper is more conducive to extracting the information of the previous frame than only the mask. By combining the long-term matching module with the short-term matching module, the whole network can achieve efficient video object segmentation without online fine tuning

# 0 Introduction

Video object segmentation is an important task in computer vision, which is applied in multiple aspects, such as intelligent monitoring, video editing and environment understanding of robots. Semi-supervised video object segmentation refers to segmenting the object in subsequent frames given the object label in the first frame. There are two types of clues used by semi-supervised video object segmentation. One is space clues: combining the current frame with the first frame for object segmentation. Another one is time clue: using the information of the previous frame to calculate the current frame, such as segmenting object of current frame with object mask predicted in the previous frame. According to different clues that are used, there are three ways of semi-supervised video object segmentation: method based on test, method based on propagation and method that involves both of them.

A typical test-based method is OSVOS proposed by Caelles et al [1], which converts video segmentation to pictures segmentation. This method trains models that merely focuses on overfitting of current video based on the first frame with labeled information of every video. VideoMatch method based on matching, proposed by Y.-T. Hu et al [2], conducts soft segmentation on the average similarity score of matching feature to generate the smooth results of prediction. These methods do not rely on time sequence information, so they can effectively process occlusion and drifting. But they depend on the first labeled frame, and can’t well process frames with obvious changes and similar objects.

A typical example of propagation-based method is MaskTrack proposed by Perazzi [3], which converts the video segmentation to guided examples of segmentation. It uses the forecast mask of previous frame as the guiding information of current frame for object segmentation. X. Li [4] further put forward the idea of re-propagation, which selects high-quality frames from video sequence to propagate forward and backward. As changes between frames are not apparent, forecast masking or optical flow of the previous frame can achieve good effects. But when dealing with cases like occlusion and disappearance, this method may propagate the wrong information to the next frame, thus affecting the results of segmentation.

Currently the main ways of semi-supervised video segmentation combine the above two methods, and make use of the Information of the first frame and the current frame at the same time. RGMP, proposed by S. Wug Oh[5], uses Siamese network to encode the features of the current frame and the first frame. It is divided into two paths on input. One path adds the mask of previous frame to current frame. Another path adds corresponding mask to the first frame. As for the features of the two paths, RGMP merely conducts superposition with no other operation. FAVOS, proposed by J. Cheng [6], divides objects of the first frame to multiple parts. For example, a person is divided into head, body and limbs. Then, it keeps track of these parts in subsequent frames and generates segmentation mask based on segmentation network of interested region. Lastly, it calculates the feature distance of segmented parts and the first frame to aggregate the parts. OSMN network, proposed by L.Yang[7], designs a modulator, inputs the target location of first frame and previous frame to network to obtain visual modulation parameters and spatial modulation parameters. Visual modulation parameters serve as weight and spatial modulation parameters serve as offset to guide current frame features so that they focus on fixed objects

Online training is an important way to enhance the performance of semi-supervised video objective segmentation. It means that after the model is trained, use the label of the first frame for each individual video to train for tens of seconds or even minutes. The longer the training time, the better the effect.. This method is developed on the basis of Lucid Data Dreaming for synthetic video frames, proposed by A. Khoreva[8]

The above methods usually make use of masking or optical flows to propagate the Information of previous frame. Sometimes online training is needed as well. Masking only represents the shapes and positions of objectives in the previous frames. In comparison, optical flow calculation needs to add optical flow detection network, which is quite complex and difficult to achieve end-to-end training. Even though online training can enhance the effects of segmentation, it takes much longer time. For this reason, this method cannot complete tasks that require higher computational efficiency. In order to enhance the real time of model calculation and make better use of time sequence information in videos, this paper puts forward long-short-term similarity matching for video object segmentation model. It uses long-term matching module to resolves issues like occlusion, disappearing and correct errors. Meanwhile, it uses short-term matching module to propagate object features. The method of this paper is an efficient model of end-to-end semi-supervised video object segmentation. Without online fine tuning, it achieves 86.5% of region similarity and 77.4% of contour accuracy in DAVIS2016 and DAVIS2017, and 21 frames per second are segmented.

This paper makes contributions in the following aspects.

1) It puts forward a new long-term matching and short-term matching convolutional neural network structure, which propagate information of previous frame to current frame through pixel-level calculation.

2) Experiment verifies that long-term matching module and short-term matching module are complementary. The combination can effectively increase the accuracy of segmentation.

3) It designs a rapid model of end-to-end semi-supervised video object segmentation, and achieves good test results on DIVIS data set.

# 1 Methods

Given object segmentation mask of the first frame, the paper designs a model of long-short-term similarity matching for video object segmentation. Long-term matching refers to mask matching between current frame and the first labeled frame. Short-term matching refers to the matching of forecast results between current frame and the first frame

## 1.1 network architecture

Network structure is shown in Fig.1 The method of this paper includes four sections: encoding modules for extracting characteristics, long-term matching modules that make use of information of the first frame,short-term matching modules that make use of information of the previous frame,and decoding modules for segmentation mask.

To go beyond the limitation of convolutional fixed reception field, and resolve the issue of deformation of nonrigid objects in movement, the paper introduces anisotropic convolution modules (AIC) of Literature [9] into the model, converting the 3D structure to 2D-AIC for processing single-frame videos. After encoders extract the features, they pass the two branches of 2D-AIC to obtain global features for long-term matching and local features for short-term matching. Then, carry out correlation operation on every pixel feature and key frame of global features of current frame to obtain global similarity map. Next, local similarity map is obtained by operating every pixel feature of local features in current frame and pixel features in corresponding areas of the previous frame. Lastly, global similarity map, local similarity map, previous frame mask and features output by encoders are transmitted to decoding module, thereby obtaining the final results through passing two refined networks. In the following sections, we will describe each module in detail:

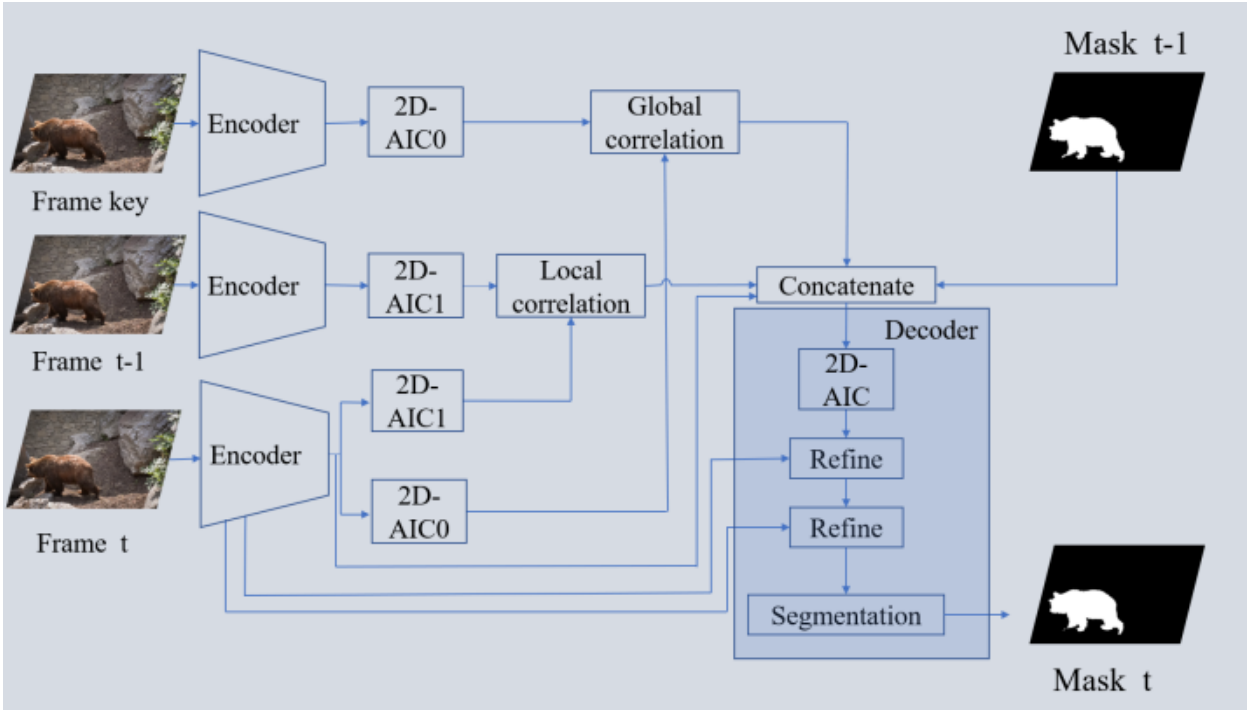

Fig 1 Network architecture.

## 1.2 Encoding module

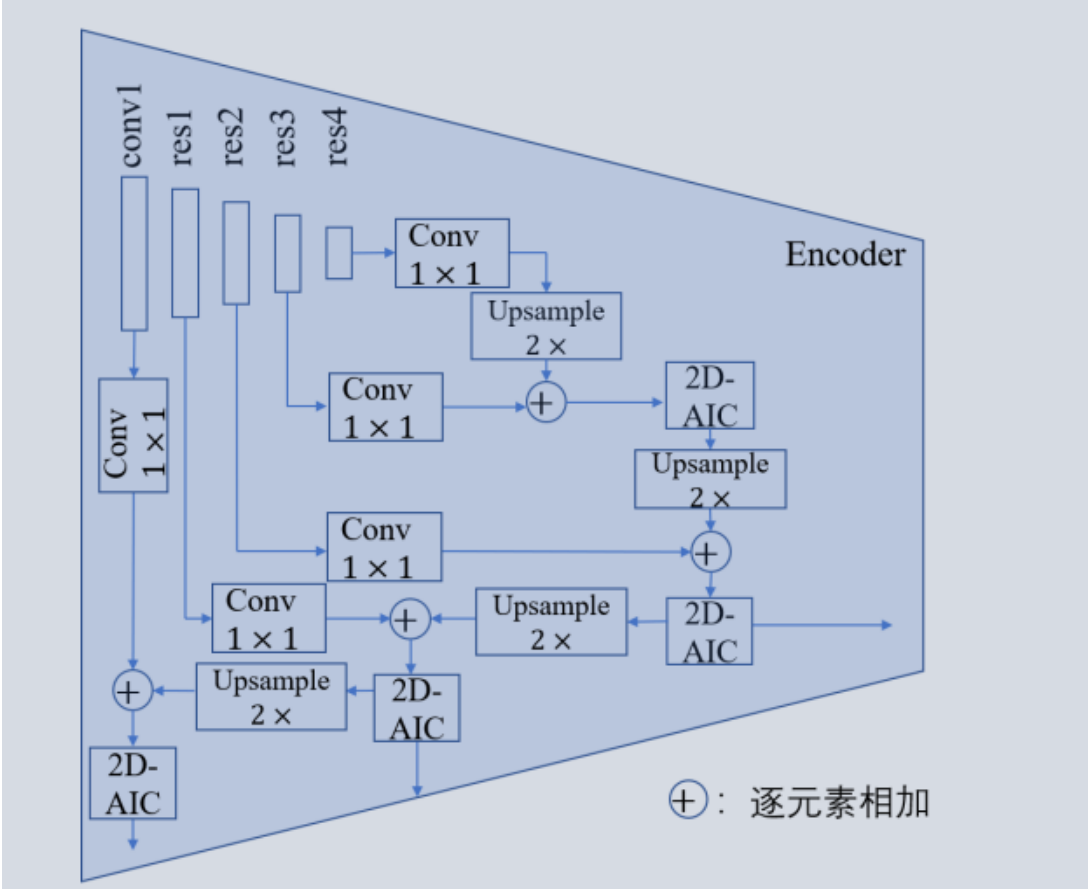

Fig 2　Encoding module

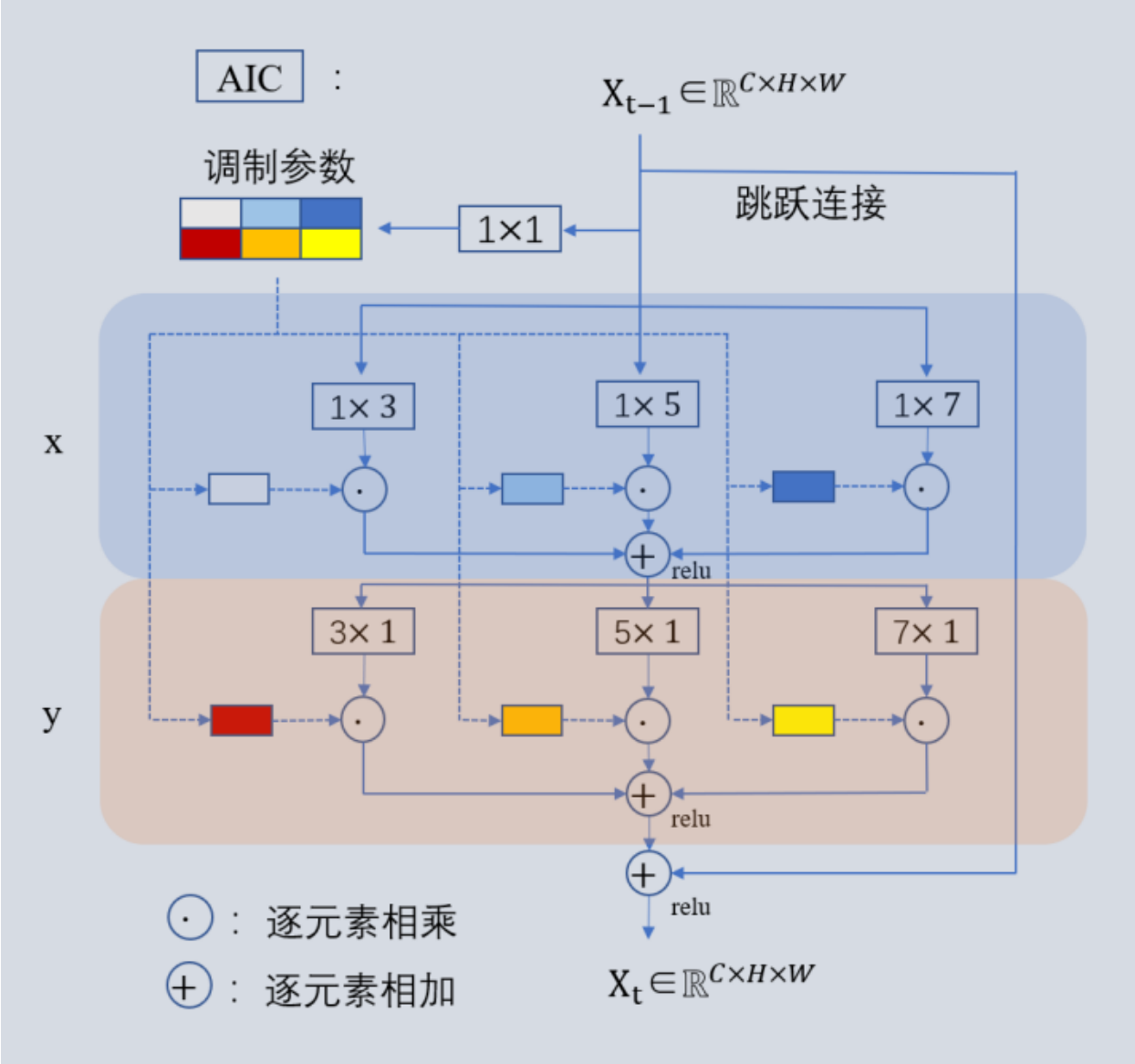

Fig 3　Anisotropic Convolution (2D-AIC)

Encoding modules use res2net[10] as the backbone network and remove the fully connected layer. Meanwhile, to make better use of multiple scale features and offer low-level features for and subsequent refined network, structures similar to FPN[11] is used in this paper. In every layer, features of previous layer carry out twofold up-sampling and add to feature of the player after 1×1 convolutional dimensionality reduction. Then, they are transmitted to a 2D-AIC structural output. Encoders of this paper have three outputs. Res2 layer output is used to extract global features and local features. The other two features are used to provide low level-features for final refined networks. Encoding modules are shown in Fig.2 2D-AIC Structures are shown in Fig.3.

**1.3 Long-term matching**

Relevant operations are widely used in object tracking. For example, for SiamRPN [12], target area and search area determine the position of objects through correlation operations. In recent years, some algorithms also introduce correlation operations to video segmentation. For example, for RANet[13]， similarity map is forged by pixel-level approaches. Then, a small-size network gives scores to the graph and chooses Layer 256 with the highest score in segmentation. In order to make use of the information of key frame and avoid occlusion and disappearing of object information in the previous frame, this paper will connect the current frame with the first frame at pixel level, as is shown in Fig.4

As for every pixel-level feature $\boldsymbol{I}_{ij}$ of extracted from global features $\boldsymbol{I} \in \mathbf{R}^{C\times H\times W}$ ($\boldsymbol{H}$ and $\boldsymbol{W}$ equal 1/8 of original picture size) of current frame, conduct correlation operations on per pixel of global features $\boldsymbol{K} \in \mathbf{R}^{C\times H\times W}$ among key frames to obtain similar picture $\boldsymbol{S}_{ij}^{g} \in \mathbf{R}^{1\times H\times W}$. As for similar picture $\boldsymbol{S}_{ij}^{g}$, after converting its dimension to $(\boldsymbol{H}\times\boldsymbol{W})\times\mathbf{1}\times\mathbf{1}$ , multiply it by $\boldsymbol{M} \in \mathbf{R}^{(H\times W)\times 1\times 1}$, the foreground (or background) of key frames. Lastly, take the maximum $\boldsymbol{N}$ （set to be 256 in this paper) values to get the corresponding pixel $\boldsymbol{G}_{ij}$ in the global similarity figure $\boldsymbol{G} \in \mathbf{R}^{N\times H\times W}$, as is shown in Equation (1)

$$\boldsymbol{G} = \{\boldsymbol{G}_{ij} | \boldsymbol{G}_{ij} = \boldsymbol{Selector}_{N}((\boldsymbol{K} \times \boldsymbol{I}_{ij}) \cdot \boldsymbol{M})\} \qquad (1)$$

$\boldsymbol{G}_{ij}$ stands for pixel feature in global similarity map of foreground (or background).$\boldsymbol{K}$ stands for feature of key frame. $\boldsymbol{I}_{ij}$ stands for pixel feature of current frame.$\boldsymbol{M}$ represents truth value of foreground (or background) in key frame. X stands for multiplication of vectors. · represents multiplication of per pixel. $\boldsymbol{Selector}_{N}$ represents the selection of maximum $\boldsymbol{N}$ values. The visualization of global similarity map is shown in Fig.5. It demonstrates the maximum response from target object in foreground and minimum response from target object in background.

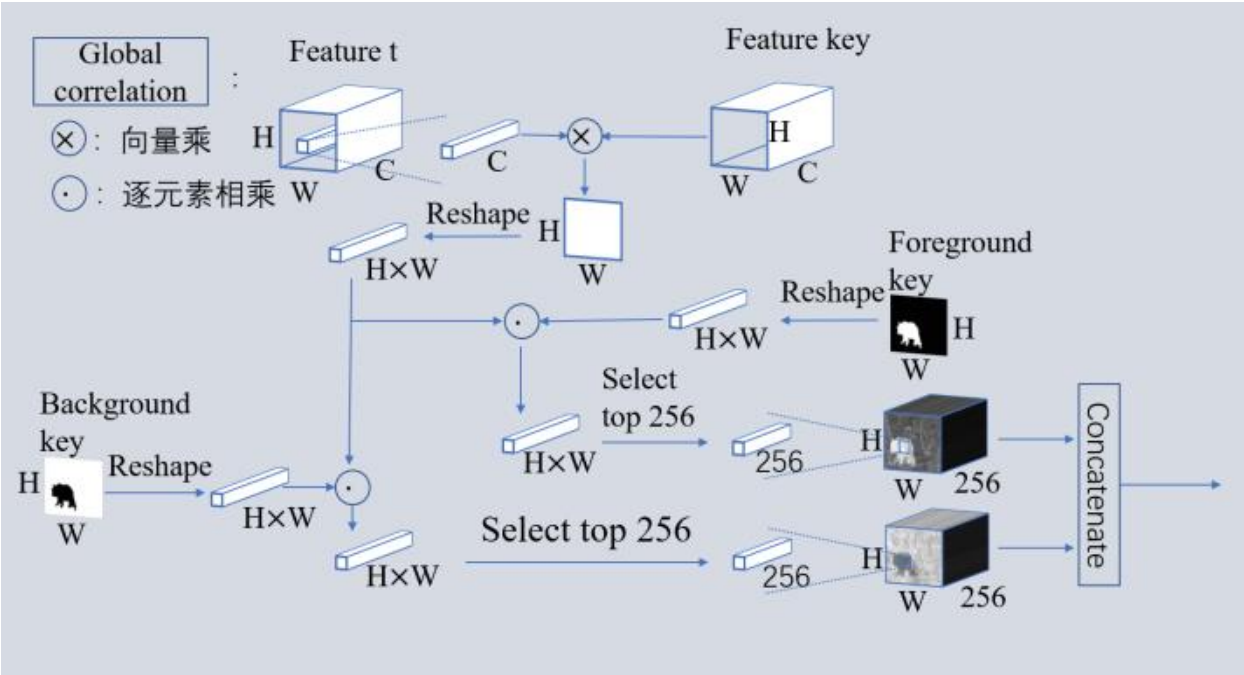

Fig 4 Long-term matching operation

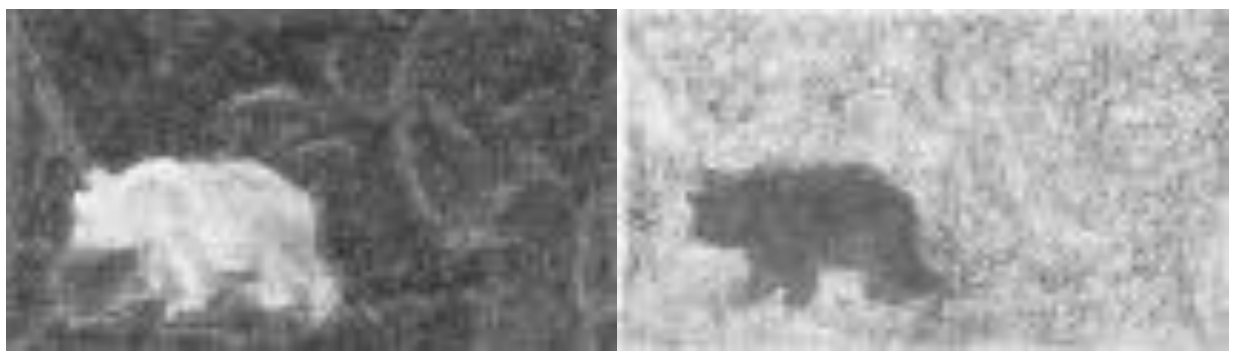

(a) (b)

Fig 5 Global similarity map

((a)foreground；(b) background)

### 1.4 short-term matching

Sequence propagation stemmed from MaskTrack[3], which also produced goods effects in other methods. But these methods merely make use of mask and optical flow of previous frame, and transfer it to network for segmentation. Mask simply reveals the position and shape of object in previous frame while ignores the object feature of previous frame. In contrast, optical flow calculation needs detection network, which is rather complex and difficult to conduct end-to-end training. In fact, we can assess which pixels of current frame are foreground or background according to the forecast of previous frame. As the changes between video frames are not apparent, it is feasible to limit the range of movement for every pixel. Inspired by mutual correlation layer of Flownet2.0[14], the paper proposes short-term matching operation, which is similar to long-term matching module.

As is shown in Fig.7, as for every pixel feature $\boldsymbol{I_{ij}^{t}}$, in local feature $\boldsymbol{I^{t}} \in \mathrm{R}^{C\times H\times W}$ ($\boldsymbol{H}$ and $\boldsymbol{W}$ equal 1/8 of original picture size) of current frame, select pixel sets $\boldsymbol{I_{d}^{t-1}}$ with distance $\boldsymbol{I_{ij}^{t-1}}$ no greater than $\boldsymbol{k}$ **(set as 8 in the paper)** on x axis and y axis in local features of previous frame $\boldsymbol{I^{t-1}} \in \mathrm{R}^{C\times H\times W}$, conduct per pixel operation to obtain similarity figure $\boldsymbol{S_{ij}^{l}} \in \mathrm{R}^{1\times(2\times k+1)\times(2\times k+1)}$. Convert the dimension of similarity figure $\boldsymbol{S_{ij}^{l}}$ to $(\boldsymbol{2}\times\boldsymbol{k}+\boldsymbol{1})\string^\boldsymbol{2}\times\boldsymbol{1}\times\boldsymbol{1}$. Afterwards, multiply it by $\boldsymbol{M} \in \mathrm{R}^{(2\times k+1)\string^2\times1\times1}$ per pixel in foreground (or background) of previous frame. Then, select maximum N (set as 256 in this paper) values to obtain corresponding pixel $\boldsymbol{L_{ij}}$ in local similarity figure $\boldsymbol{L} \in \mathrm{R}^{N\times H\times W}$, as is shown in Fig.2

$$\boldsymbol{L} = \{\boldsymbol{L_{ij}} | \boldsymbol{L_{ij}} = \boldsymbol{Selector_N}((\boldsymbol{I_d^{t-1}} \times \boldsymbol{I_{ij}^{t}}) \cdot \boldsymbol{M}$$
$$(\boldsymbol{d} \in ([\boldsymbol{i}-\boldsymbol{k}, \boldsymbol{i}+\boldsymbol{k}] \cap [\boldsymbol{j}-\boldsymbol{k}, \boldsymbol{j}+\boldsymbol{k}]))\} \quad (2)$$

Among the pixel $\boldsymbol{I_{ij}^{t}}$ at （$\boldsymbol{i}$，$\boldsymbol{j}$） of current frame feature, foreground feature (or background feature) of previous frame, take （$\boldsymbol{i}$，$\boldsymbol{j}$） as the center, conduct per pixel calculation on pixel sets on x axis and y axis with distance no greater than $\boldsymbol{k}$, in order to obtain the similarity value of $(\boldsymbol{2}\times\boldsymbol{k}+\boldsymbol{1})\string^\boldsymbol{2}$. Then, select the top N values to form foreground similarity (or background similarity figure)$\boldsymbol{L} \in \mathrm{R}^{N\times H\times W}$. visualization of local similarity figure is shown in Fig. 6. in comparison with global similarity map, local similarity map gets rid of many interference factors, so the results are more explicit.

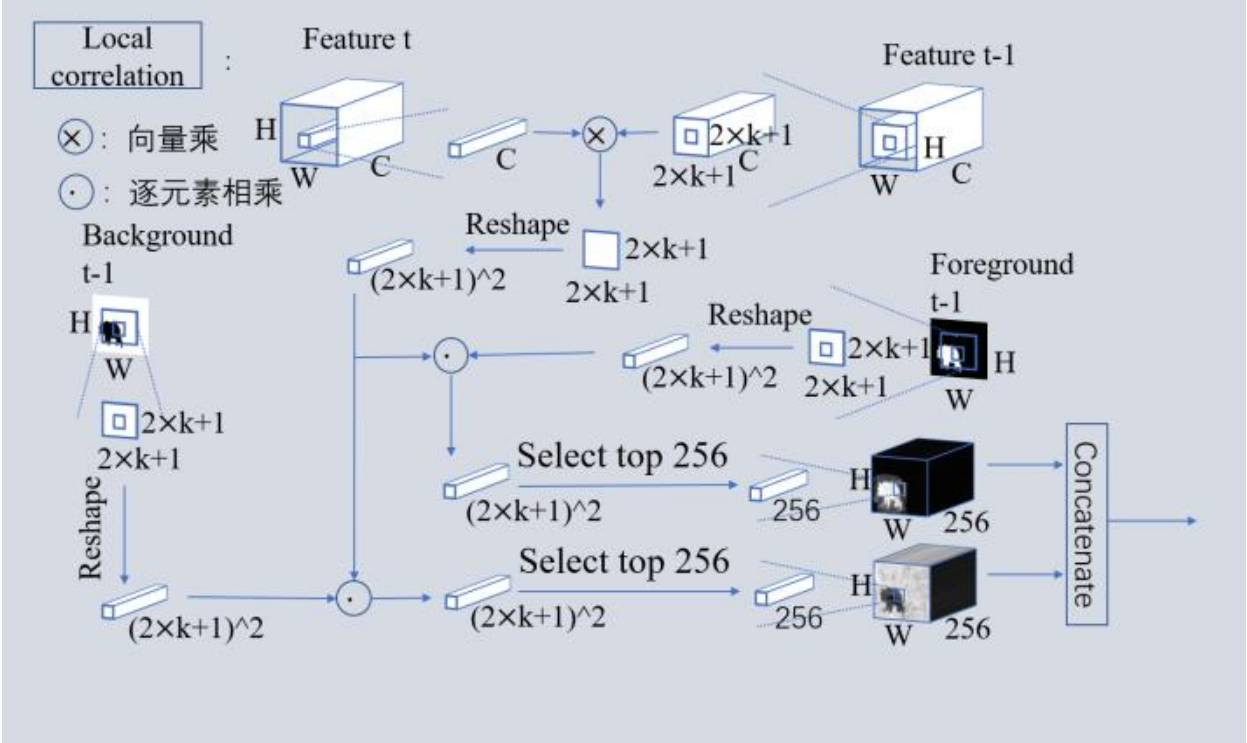

Fig 6 Short-term matching operation

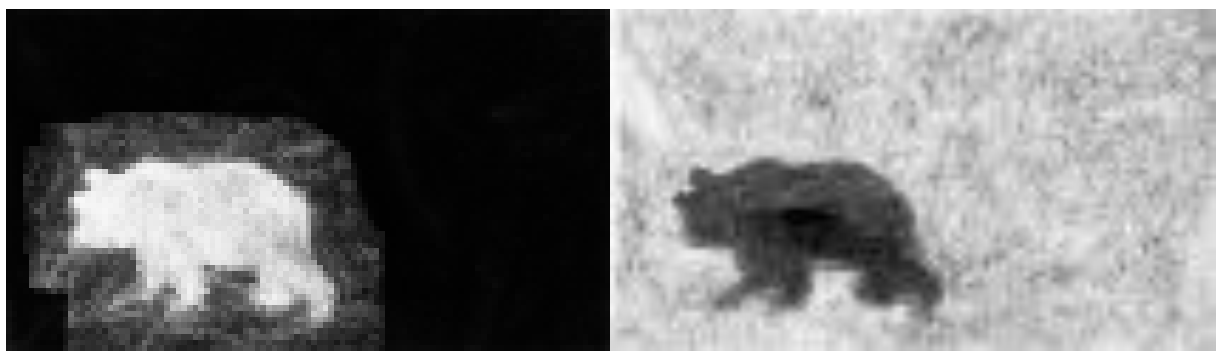

(a) (b)

Fig 7 Local similarity map

((a)foreground；(b) background)

### 1.5 Decoding module

Decoding modules include two refined networks for up-sampling and one conv $3\times3$ segmentation network for extracting probability graph of final results.

Refined network is shown as Fig.8. Feature i stands for the same-layer features output by encoders, which will be processed by 2D-AIC. Afterwards, add the features to those of previous layer that passed twofold up-sampling. Lastly, after passing two refined networks, segmentation is carried out at 1/2 size of the picture..

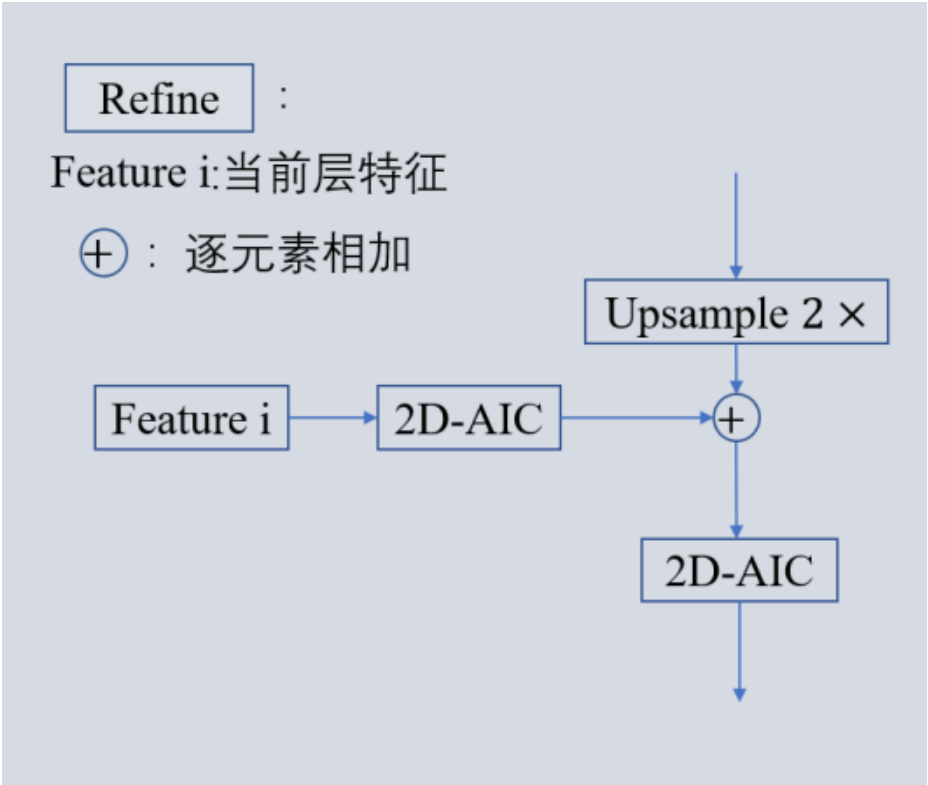

Fig 8 Refined module

# 2. Experiment

## 2.1 details of training

The network in this article is end-to-end, and the backbone is initialized with the res2net50 model parameters pre-trained on the ImageNet dataset. The optimizer is the Adam optimizer, the learning rate is set to a fixed 0.00001, and the loss function is Focal loss. The dataset uses YouTube-VOS and DAVIS, and trains 200,000 steps on 4 NVIDIA GeForce TITAN Xp with the batch size being set to 12. Data augmentation adopts random cropping, random size transformation and random flip.

## 2.2 Single-target video segmentation

Single target video separation experiment is carried out on DAVIS2016[18] data set. The data set has 50 videos, including 2079 frames of 30 training videos and 1376 frames of 20 testing videos. The first frame of every video displays the labeled Information. The evaluation index involves regional similarity J and contour accuracy F. By comparing the DAVIS2016 verification set with methods in Table 1, the solutions of this paper achieves advanced levels without online tuning. Average values *J&F* of regional similarity and contour accuracy reach 86.5%. After model training is completed, the online inference rate reaches 21 FPS on GeForce TITAN Xp computer card, achieving favorable balance in terms of time and accuracy.

Table 2 shows the results of ablation experiment. To explore the roles of network structure in various modules, only short-term module and long-term modules are retained for training, and results are shown in the first and second rows of Table 2. Both of them achieve very good results. But they are significantly different from results in the fourth row, which combine two module training. *J&F* indexes decrease by 3.7% and 2.9% respectively. Only short-term matching module may reduce the accuracy, as the lack of the first frame correction cause the result that errors transfer from the previous frame to current frame. As is shown in Fig.9, with no correction of true value from the first frame, errors increases along with the growth of frame number. Only long-term matching module may also reduce accuracy. This is because the latter targets show great differences from the first frame. The sheer matching with the first frame pixels is not sufficient to capture targets, especially when the target size changes are excessively significant, as is shown in Fig.10. Because it is excessively different from the target of first frame, the network cannot detect the target. In order to determine the contribution of mask and short-term matching module to sequence propagation, this paper gets rid of short-term matching modules and mask to carry out experiment, the result of which are shown in the 3rd and 4th rows of Table 2. *J&F* with short-term matching removed increases by 2.2% than with mask removed. It proves the validity of short-term matching module proposed in this paper. Even though short-term matching module have very good effects, it cannot sufficiently represent the shape and location of objects, so it still needs the complimentary function of mask. With the additional role of mask, this paper's method increases by 1.6% in terms of *J&F* index.

**Table 1 Results of different methods on the DAVIS 2016 validation set**

| Method | *J&F*↑ | *J* mean↑ | *J* recall↑ | *J* decay↓ | *F* mean↑ | *F* recall↑ | *F* decay↓ | FPS↑ |
|---|---|---|---|---|---|---|---|---|
| OSVOS*[1] | 80.2 | 79.8 | 93.6 | 14.9 | 80.6 | 92.6 | 15.0 | - |
| MaskTrack*[3] | 77.6 | 79.7 | 93.1 | 8.9 | 75.4 | 87.1 | 9.0 | - |
| PReMVOS*[20] | 86.8 | 84.9 | 96.1 | 8.8 | 88.6 | 94.7 | 9.8 | - |
| VideoMatch[2] | - | 81.0 | - | - | - | - | - | - |
| RGMP[5] | 81.8 | 81.5 | 91.7 | 10.9 | 82.0 | 90.8 | 10.1 | 7.7 |
| FAVOS[6] | 81.0 | 82.4 | 96.5 | 4.5 | 79.5 | 89.4 | 5.5 | 0.56 |
| OSMN[7] | 73.5 | 74.0 | 87.6 | 9.0 | 72.9 | 84.0 | 10.6 | 7.7 |
| RANet[13] | 85.5 | 85.5 | 97.2 | 6.2 | 85.4 | 94.9 | 5.1 | 30.3 |
| Ours | 86.5 | 85.7 | 97.1 | 5.1 | 87.3 | 96.1 | 4.9 | 21.3 |

Note: Red shows the optimum value of each row. Green shows less optimum value of each row. * means online training. ↑ means the higher the better. ↓ means the lower the better.

**Table 2　Ablation study results**

| Strategy. | *J&F*↑ | *J* mean↑ | *J* recall↑ | *J* decay↓ | *F* mean↑ | *F* recall↑ | *F* decay↓ |
|---|---|---|---|---|---|---|---|
| Long-term matching modules and mask removed | 81.2 | 80.6 | 91.4 | 7.4 | 81.8 | 90.9 | 6.7 |
| Short-term matching modules and mask removed | 82.0 | 81.0 | 92.7 | 11.3 | 83.0 | 93.1 | 11.7 |
| Short-term matching modules removed | 82.7 | 81.5 | 93.1 | 9.0 | 83.9 | 93.3 | 10.6 |
| Mask removed. | 84.9 | 83.9 | 94.7 | 6.0 | 85.9 | 94.4 | 6.9 |
| Methods of this paper. | 86.5 | 85.7 | 97.1 | 5.1 | 87.3 | 96.1 | 4.9 |

Note: Red shows the optimum value of each row. Green shows less optimum value of each row. * means online training. ↑ means the higher the better. ↓ means the lower the better.

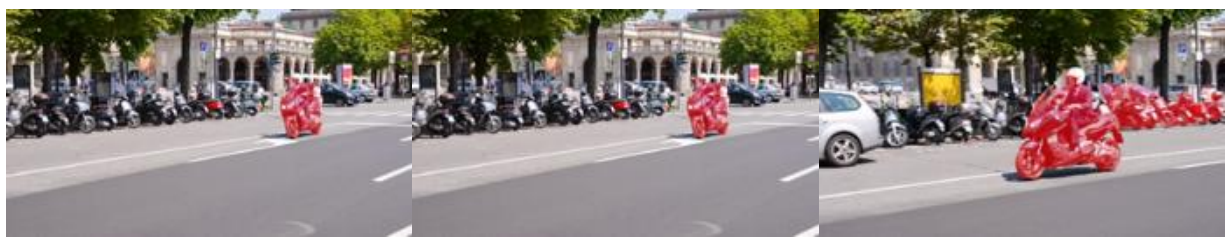

Frame 2　　Frame 12　　Frame 22

Fig 9　Error propagation

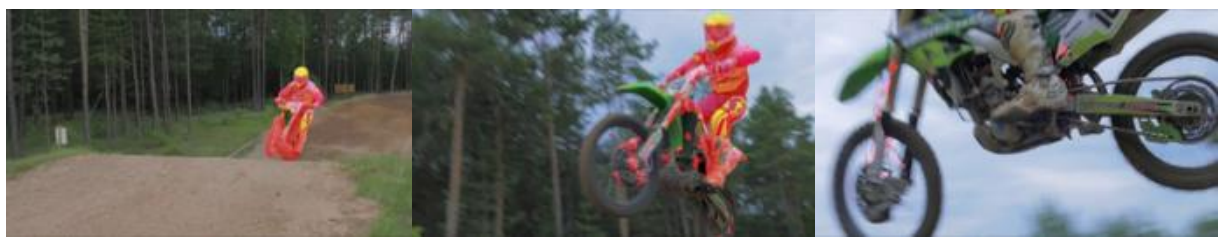

Frame 2　　Frame 12　　Frame 22

Fig 10　Error match

## 2.3　Multiple object video segmentation

In multiple object video segmentation, many similar targets are very likely to shield or miss each other, which makes this process challenging. In this paper, multiple object video segmentation is considered as single object video segmentation. After all the targets are segmented from every frame, select the category with the maximum probability. The reason is that encoding modules, long-term matching modules and short-term matching modules are relatively time-consuming. In every frame, the multiple targets can be shared commonly, which consumes approximately 42 milliseconds. Encoding module is the only thing linearly dependent on target numbers, and each target consumes approximately 11 milliseconds. Therefore, the algorithm of this paper still achieves high efficiency in terms of multiple object video segmentation. The experimental results of verification set and testing set on multiple object data set DAVIS2017[19] are shown in Table 3 and 4. It can be seen the method of this paper still produces favorable results in multiple objects. *J&F* indexes of PReMVOS[20] increase by 0.5% and 4.2% than the method of this paper. This is because it utilizes the method of online training. For every video, the training of the first frame needs to last dozens of seconds and even several minutes. In comparison, the method of this paper can achieve equivalent results without the need for online training.

**Table 3　Results of different methods on the DAVIS 2017 validation set**

| Method | *J&F*↑ | *J* mean↑ | *J* recall↑ | *J* decay↓ | *F* mean↑ | *F* recall↑ | *F* decay↓ |
|---|---|---|---|---|---|---|---|
| OSVOS* | 60.3 | 56.6 | 63.8 | 26.1 | 63.9 | 73.8 | 27.0 |
| PReMVOS* | 77.9 | 73.9 | 83.1 | 16.2 | 81.8 | 88.9 | 19.5 |
| VideoMatch | - | 56.5 | - | - | - | - | - |
| RGMP | 66.7 | **64.8** | 74.1 | 18.9 | 68.6 | 77.7 | 19.6 |
| FAVOS | 58.2 | 54.6 | 61.1 | **14.1** | 61.8 | 72.3 | **18.0** |
| OSMN | 54.8 | 52.5 | 60.9 | 21.5 | 57.1 | 66.1 | 24.3 |
| RANet | 65.7 | 63.2 | 73.7 | 18.6 | 68.2 | 78.8 | 19.7 |
| Ours | **77.4** | **73.9** | **83.6** | **12.9** | **80.8** | **91.3** | **15.7** |

Note: Red shows the optimum value of each row. Green shows less optimum value of each row. * means online training. ↑ means the higher the better. ↓ means the lower the better.

**Table 4 Results of different methods on the DAVIS 2017 test set**

| Method | *J&F*↑ | *J* mean↑ | *J* recall↑ | *J* decay↓ | *F* mean↑ | *F* recall↑ | *F* decay↓ |
|---|---|---|---|---|---|---|---|
| OSVOS* | 50.9 | 47.0 | 52.1 | 19.2 | 54.8 | 59.7 | 19.8 |
| PReMVOS* | **71.6** | **67.5** | **76.8** | 21.7 | **75.8** | **84.3** | 20.6 |
| RGMP | 52.8 | 51.3 | 59.0 | 34.3 | 54.4 | 61.9 | 37.2 |
| FAVOS | 43.6 | 42.9 | 48.1 | **18.1** | 44.2 | 51.1 | 19.8 |
| OSMN | 41.3 | 37.7 | 38.9 | 19.0 | 44.9 | 47.4 | **17.4** |
| RANet | 55.4 | 53.4 | 61.9 | 21.9 | 57.3 | 67.7 | 22.1 |
| Ours | **67.4** | **63.7** | **72.7** | **16.9** | **71.2** | **81.4** | **16.5** |

Note: Red shows the optimum value of each row. Green shows less optimum value of each row. * means online training. ↑ means the higher the better. ↓ means the lower the better.

### 2.4 qualitative results

Fig. 11 shows the segmentation results of every 10 frames of four videos. The first video and second video are respectively about horse riding and race car drifting. These two videos verify the processing capability of the algorithm about deformation and rapid movement. The 3rd video and 4th video are respective concerned with five goldfish in the ocean and three pedestrians in the crowd. It can be seen that despite the occlusion of similar targets and complicated background, algorithms of this paper still produce favorable results. More results of video segmentation can be seen in the following websites:

https://www.bilibili.com/video/BV1jK4y1Y7yd/
https://www.bilibili.com/video/BV1MC4y1t7R2/
https://www.bilibili.com/video/BV1Bh411d72y/

## 3. Conclusion

In order to deal with under-utilization of previous frame information, this paper puts forward short-term matching modules, whose the favorable effects are verified by experiments. By combining long-term matching module with short-term matching module, a simple and fast method of end-to-end video segmentation is proposed in this paper. It has favorable capability of processing in the following scenarios: target occlusion, deformation, rapid movement. Compared with other methods, the network structure of this paper achieves competitive results in terms of DAVIS2016 and DAVIS2017. Besides, without the need for online training, the speed of segmentation is enhanced significantly. Based on the achievements of this paper, subsequent research and studies will make further use of prediction results of previous frames to segment the objects of current frames.

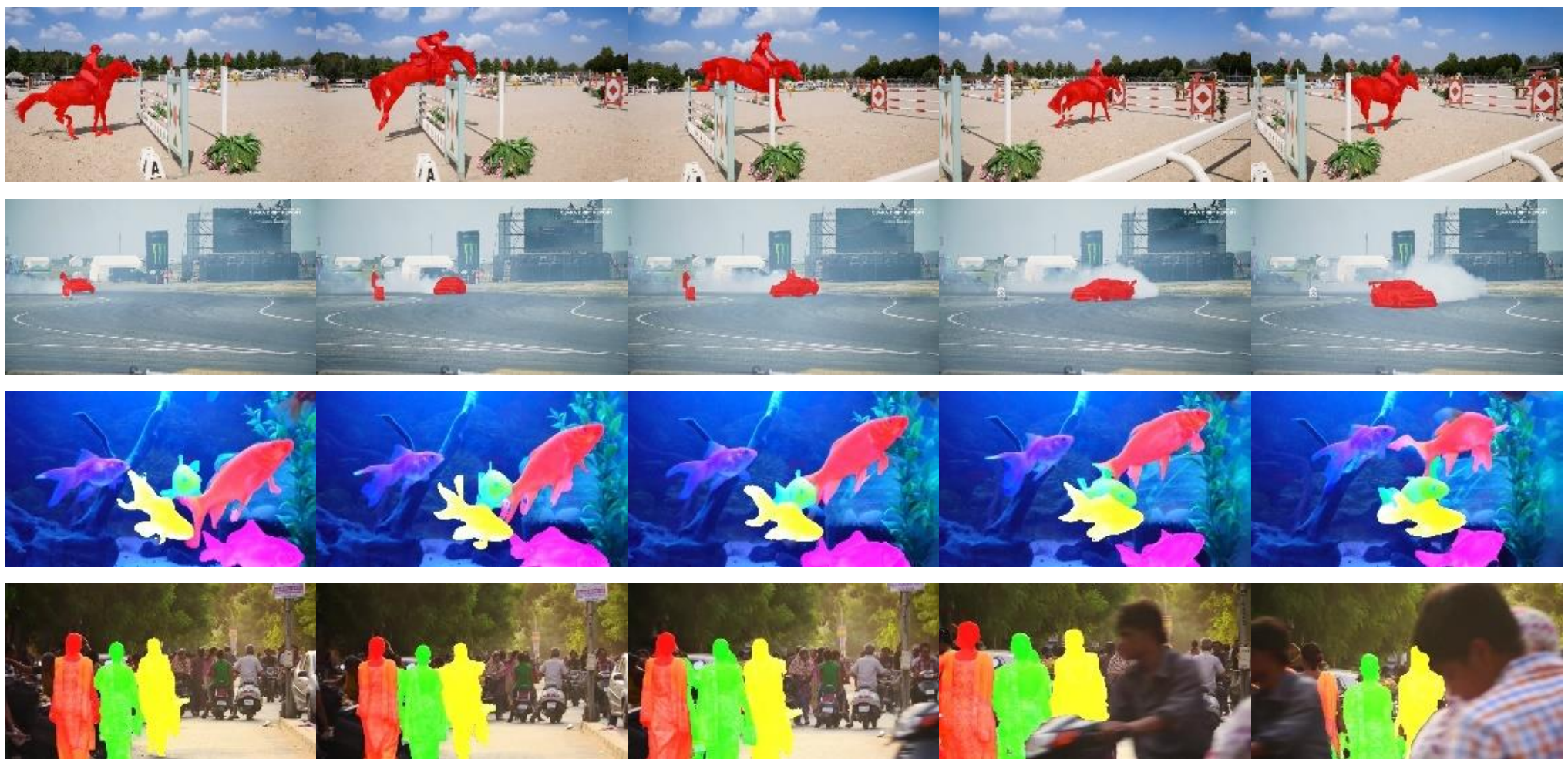

Frame 2 Frame 12 Frame 22 Frame 32 Frame 42

Fig 11 Qualitive results